\documentclass[letterpaper]{article}
\usepackage{aaai17}
\usepackage{times}
\usepackage{helvet}
\usepackage{courier}
\usepackage{graphicx}
\usepackage{comment}
\usepackage{url}      
\usepackage{tabularx}
\usepackage{amssymb}
\usepackage{booktabs}
\usepackage{balance}  
\usepackage{wasysym}

\begin{document}
%
\title{Long-Term Trends in the Public Perception of Artificial Intelligence}
\author{Ethan Fast \and Eric Horvitz\\
  {\tt ethaen@stanford.edu, horvitz@microsoft.com}}

\maketitle
\begin{abstract}
Analyses of text corpora over time can reveal trends in beliefs, interest, and sentiment about a topic. We focus on views expressed about artificial intelligence (AI) in the New York Times over a 30-year period. General interest, awareness, and discussion about AI has waxed and waned since the field was founded in 1956. We present a set  of measures that captures levels of engagement, measures of pessimism and optimism, the prevalence of specific hopes and concerns, and topics that are linked to discussions about AI over decades. We find that discussion of AI has increased sharply since 2009, and that these discussions have been consistently more optimistic than pessimistic. However, when we examine specific concerns, we find that worries of loss of control of AI, ethical concerns for AI, and the negative impact of AI on work have grown in recent years. We also find that hopes for AI in healthcare and education have increased over time.
\end{abstract}

\section{Introduction}

\begin{quote}
\footnotesize
Artificial intelligence will spur innovation and create opportunities, both for individuals and entrepreneurial companies, just as the Internet has led to new businesses like Google and new forms of communication like blogs and social networking. Smart machines, experts predict, will someday  tutor students, assist surgeons and safely drive cars.

\textit{Computers Learn to Listen, and Some Talk Back}. NYT, 2010

In the wake of recent technological advances in computer vision, speech recognition and robotics, scientists say they are increasingly concerned that artificial intelligence technologies may permanently displace human workers, roboticize warfare and make Orwellian surveillance techniques easier to develop, among other disastrous effects.

\textit{Study to Examine Effects of Artificial Intelligence}. NYT, 2014

\normalsize
\end{quote}

These two excerpts from articles in the New York Times lay out competing visions for the future of artificial intelligence (AI) in our society. The first excerpt is optimistic about the future of AI---the field will ``spur innovation,'' creating machines that tutor students or assist surgeons---while the second is pessimistic, raising concerns about displaced workers and dystopian surveillance technologies. But which vision is more common in the public imagination, and how have these visions evolved over time? 

Understanding public concerns about AI is important, as these concerns can translate into regulatory activity with potentially serious repercussions \cite{ai100}. For example, some have recently suggested that the government should regulate AI development to prevent existential threats to humanity \cite{elon-regulatory}. Others have argued that racial profiling is implicit in some machine learning algorithms, in violation of current law \cite{machine-bias}. More broadly, if public expectations diverge too far from what is possible, we may court the smashed hopes that often follow from intense enthusiasm and high expectations.   

AI presents a difficult case for studies of topic sentiment over time because the term is not precisely defined. Lay people and experts alike have varied understandings of what ``artificial intelligence'' means \cite{ai100}.  Even in the narrowest, engineering-centric definitions, AI refers to a broad constellation of computing technologies.

We present a characterization of impressions expressed about AI in the news over 30 years. First, we define a set of indicators that capture levels of engagement, general sentiment, and hopes and concerns about AI. We then apply and study these indicators across 30 years of articles from the New York Times. As a proxy for public opinion and engagement, no other corpus extends so far into the past to capture how a general audience thinks about AI. Moving forward, we can apply these indicators to present day articles in an ongoing effort to track public perception.

Our study relies on a combination of crowdsourcing and natural language processing. For each article under analysis, we extract all mentions of artificial intelligence and use paid crowdsourcing to annotate these mentions with measures of relevance, their levels of pessimism or optimism about AI, and the presence of specific hopes and concerns, such as ``losing control of AI'' or ``AI will improve healthcare.'' These annotations form the basis of the indicators and allow us to bootstrap a classifier that can automatically extract impressions about AI, with applications to tracking trends in new articles as they are generated.

To study how public perception of AI has changed over time, we analyze the set of indicators for articles published in the New York Times between January 1986 and June 2016. We address four research questions:

\textbf{R1}: How prominent is AI in the public discussion today, as compared to the past? 

\textbf{R2}: Have news articles become generally more optimistic or more pessimistic about AI over time? 

\textbf{R3}: What ideas are most associated with AI over time? 

\textbf{R4}: What specific ideas were the public concerned about in the past, which are no longer concerns today? Likewise, what new ideas have arisen as concerns?

When we examine the impression indicators across historical data, we find that AI has generally taken on a stronger role in public discussion over time---with a few notable blips, such as the so-called \textit{AI winter} in 1987. Further, we find that the mood of discussion has generally remained more optimistic over time, although this trend is not common across all concerns (e.g., AI's impact on work). Finally, we discover that some ideas, such as ``AI for healthcare'' or ``losing control of AI,'' are more common today than in the past. Other ideas, for example, that ``AI is not making enough progress'' or that ``AI will have a positive impact on work,'' were more common in the past than they are today. 

\section{Indicators of Impressions about AI}

We capture the intensity of engagement on AI in the news as well as the prevalence of a diverse set of hopes and concerns about the future of AI. We took inspiration from the Asilomar Study of 2008-09 \cite{asilomar} and the One Hundred Year Study on Artificial Intelligence \cite{100years} to create measures that capture a long-term perspective for how AI impacts society. 

\subsection{General Measures}
We have included the following general measures:

\textbf{Engagement}. This measure serves as a proxy for public interest and engagement around AI, capturing how much AI is discussed in the news over time.

\textbf{Optimism vs. Pessimism}. This measure captures the attitude of a discussion---the degree to which it implies a sense of optimism or pessimism about the future of AI. This attitude can stem from technological progress, such as optimistic reporting on new breakthroughs in deep learning. But it can also be influenced by the impact of new technologies on society: for example, the time-saving benefits of a self-driving car (a form of optimism); or the dangers of surveillance as data is collected and mined to track our leanings, locations, and daily habits (a form of pessimism). We include such \textit{attitudinal} leanings as an indicator to track these high-level trends. Notably, traditional sentiment analysis does not capture optimism versus pessimism. 

The remainder of our indicators capture common hopes and concerns about the future of AI.

\subsection{Hopes for Artificial Intelligence}

We have included the following hopes for AI as indicators:

\textbf{Impact on work (positive)}: AI makes human work easier or frees us from needing to work at all, e.g., by managing our schedules, automating chores via robots.

\textbf{Education}: AI improves how students learn, e.g., through automatic tutoring or grading, or providing other kinds of personalized analytics.

\textbf{Transportation}: AI enables new forms of transportation, e.g., self-driving cars, or advanced space travel.

\textbf{Healthcare}: AI enhances the health and well-being of people, e.g., by assisting with diagnosis, drug discovery, or enabling personalized medicine.

\textbf{Decision making}: AI or expert systems help us make better decisions, e.g., when to take a meeting, or case-based reasoning for business executives. 

\textbf{Entertainment}: AI brings us joy through entertainment, e.g., though smarter enemies in video games.

\textbf{Singularity (positive)}: A potential singularity will bring positive benefits to humanity, e.g., immortality.

\textbf{Merging of human and AI (positive)}: Humans merge with AI in a positive way, e.g., robotic limbs for the disabled, positive discussions about potential rise of transhumanism.

\subsection{Concerns for Artificial Intelligence}

We have also considered the following concerns for AI:

\textbf{Loss of control}: Humans lose control of powerful AI systems, e.g., Skynet or ``Ex Machina'' scenarios.

\textbf{Impact on work (negative)}: AI displaces human jobs, e.g., large-scale loss of jobs by blue collar workers.

\textbf{Military applications}: AI kills people or leads to instabilities and warfare through military applications, e.g., robotic soldiers, killer drones.

\textbf{Absence of Appropriate Ethics}: AI lacks ethical reasoning, leading to negative outcomes, e.g., loss of human life.

\textbf{Lack of progress}: The field of AI is advancing more slowly than expected, e.g., unmet expectations like those that led to an AI Winter. 

\textbf{Singularity (negative)}: The singularity harms humanity, e.g., humans are replaced or killed.

\textbf{Merging of human and AI (negative)}: Humans merge with AI in a negative way, e.g., cyborg soldiers.

\section{Data: Thirty Years of News Articles}

We conduct our analysis over the full set of articles published by the New York Times between January 1986 and May 2016---more than 3 million articles in total. 

We have created this dataset by querying the New York Times public API for metadata (e.g., title of article, section of paper, current URL) associated with articles published on each individual day within the scope of our analysis. For each article, we then scrape the full text from its URL using the \textit{BeautifulSoup} python package.

Next, we annotate articles on AI. Unfortunately, crowdsourcing annotations for full news articles is a complex task, requiring a large time expenditure for workers. For this reason we segment our data into paragraphs. In news articles, paragraphs tend to be self-contained enough that workers can annotate them accurately without reading the rest of the article. 
This makes them a good middle ground between full documents and individual sentences. For example:

\begin{quote}
\small
Artificial intelligence ``has great potential to benefit humanity in many ways.'' An association with weaponry, though, could set off a backlash that curtails \textbf{its} advancement.
\normalsize
\end{quote}

While the above paragraph clearly discusses AI for military applications, annotating the same text at the sentence or document level might not produce that annotation. For example, sentence level annotations would not connect ``its'' with AI, and document level annotations would too often result in workers missing the relevant passage, but paragraph level annotations easily capture this relationship.

It is expensive to crowdsource annotations the tens of millions of paragraphs in the dataset, so we filter these paragraphs to the set that contain ``artificial intelligence'', ``AI'', or ``robot''. (We include ``robot'' to increase coverage---we are not concerned with false positives at this stage, as we will later filter for relevance.) In total, we retrieve more than 8000 paragraphs that mention AI over a thirty year period.

\section{Crowdsourcing to Annotate Indicators}

Crowdsourcing provides an efficient way to gather annotations for our dataset of AI-related paragraphs. In this section, we present the details of the approach.

\subsection{Task Setup}

We used Amazon Mechanical Turk (AMT) to collect annotations for the more than 8000 AI-related paragraphs in our dataset. We assigned each paragraph to a task with multiple components. First, we collected annotations for \textit{attitude} about the future of AI (from pessimistic to optimistic) on a 5-point Likert scale. We then collected low level annotations for all of the specific \textit{hopes and concerns} developed. We requested binary labels that indicate whether the hope or concern is present in the paragraph (e.g., AI will have a negative impact on work). 
Finally, to ensure that unrelated paragraphs do not bias our results, we collected high-level annotations for \textit{AI relevance} (from strongly unrelated to strongly related) on a 5-point Likert scale.

We assigned each AMT task to three independent workers in pursuit of reliable labels \cite{get-another-label}. We provided examples to better ground the task \cite{examples}, and recruited Masters workers to ensure quality results. We paid \$0.15 per task in line with guidelines for ethical research \cite{dynamo}, for a total cost of \$3825. In Supplementary Material, we include a template for the task we used.

In general, workers show high rates of agreement over labels for AI relevance and mood. Across paragraphs, 97\% of workers agreed that they were either at least somewhat related to AI or else unrelated. 70\% of workers agreed when distinguishing between optimistic and pessimistic articles. When interpreting ratings for AI relevance and attitude, we take the average across workers.

\subsection{Interpreting annotations for hopes and concerns}

One decision we must make is how to interpret the crowd annotations for  hopes and concerns. Should we require that all three workers mark a paragraph with a hope or concern to include it in our data? Or trust the majority vote of two workers? Or require only one worker's vote? Requiring a larger number of votes will reduce the rate of false positives (i.e., labeling a paragraph with a hope or concern it does not exhibit), but may increase the rate of false negatives. 

To determine the best approach, we established a ground truth dataset for one example concern, \textit{military applications} for  AI. We examined each paragraph associated with this concern by at least one worker and determined whether it in fact expressed that concern. This allowed us to calculate precision and
a proxy for recall\footnote{This number will be higher than true recall, but describes how many true positives we miss in the subset of ground truth data.} across voting schemes.

We find a trade-off between precision and recall (Table \ref{tbl:rates}). Requiring two or more votes results in precision of 100\%, but recall of 59\%. Alternatively, only requiring one worker vote results in precision of 80\% and recall of 100\%. In light of these numbers and the fact that many of our hopes and concerns are covered sparsely in the dataset (for example, we see only 231 mentions of ``loss of control of AI'' across the thirty year corpus), we require only one worker vote to label a paragraph with a given hope or concern.

\begin{table}[!ht]
  \renewcommand{\arraystretch}{1.3}
  \begin{tabular}{p{8.5em}p{3em}p{3em}p{3em}}
   \textbf{\# worker votes} & 1 & 2 & 3\\
   \hline
   \textbf{precision} & 0.81 & 1.00 & 1.00 \\
   \textbf{recall} & 1.00 & 0.59 & 0.18 \\

  \end{tabular}
  \caption{How voting schemes impact a paragraph's association with \textit{military applications} in terms of precision and a proxy for recall on ground truth data. 
  }
  \label{tbl:rates}
\end{table}

\section{Trends in the Public Perception of AI}

Using our crowdsourced annotations, we now analyze trends in public impressions of AI over 30 years of news articles. We conduct this analysis through four research questions.

\begin{figure}[!t]
\centering
\includegraphics[width=1.0\columnwidth]{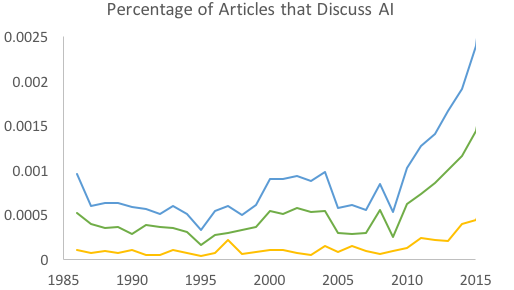}
\caption{Articles that discuss AI over time, as a percentage of the total number of articles published per year. The green line plots \textit{optimistic} articles and the yellow line plots \textit{pessimistic} articles. AI discussion has exploded since 2009, but levels of pessimism and optimism have remained balanced.}
\label{fig:general}
\end{figure}

\subsection{R1: How prominent is discussion of AI?}

A natural starting point for understanding the public perception of AI is to examine how frequently it is discussed over time, and what events influence this discussion. 

We capture this idea through an \textit{engagement} measure. To compute this, we first filter the data to include only paragraphs with an average AI relevance rating of more than 3.0, as determined by our crowdsourcing pipeline. We then aggregate these paragraphs by the news article they appear in, and further aggregate articles by their year of publication. This leaves us with data that count how many articles that mention AI are published every year from 1986 to 2016. Finally, we normalize these counts by the total volume of articles published each year. 

We present a graph of AI engagement in Figure \ref{fig:general}. Most strikingly, we observe a dramatic rise in articles that mention AI beginning in late 2009.  While the cause of this rise is unclear, it occurs following a renaissance in the use of neural nets ("deep learning") in natural language and perceptual applications, and after a front page story discussed the Asilomar meeting \cite{asilomar}. We also observe a fall in AI discussion that corresponds with the start of the 1987 AI winter---reaching its lowest level in 1995. 

\subsection{R2: Have impressions reported in the news become more optimistic or pessimistic about AI?}

In addition to engagement, we studied indicators for levels of pessimism and optimism in the coverage of AI. How have these levels changed over time? While it is easy to imagine a public celebration of AI technology as it becomes more common, it is also possible to imagine greater levels of concern, as people worry about changes they cannot control.

To track public sentiment over time, we draw on the dataset of AI-related news articles, aggregated by year, that we created for R1. We divide each year into counts of optimistic and pessimistic articles, as determined by the \textit{attitude} rating in our crowdsourcing pipeline (considering an article optimistic if it has an average rating greater than 3, and pessimistic if it has an average rating of less than three). 

We present the resulting trends in Figure \ref{fig:general}. In general, AI has had consistently more optimistic than pessimistic coverage over time, roughly 2-3 times more over the 30 year period. Since 2009, both optimistic and pessimistic coverage have exploded along with general interest in AI.

\begin{figure*}[!t]
\centering
\includegraphics[width=2.0\columnwidth]{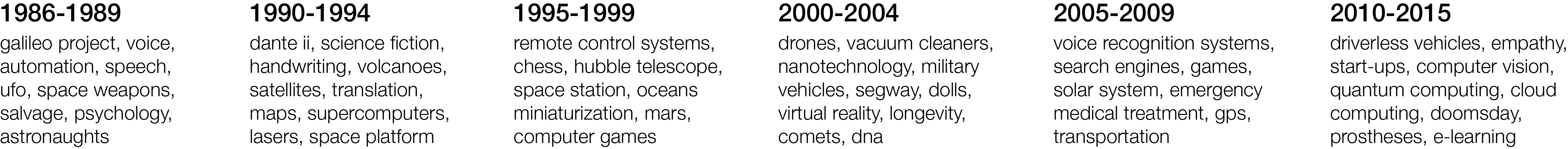}
\caption{New York Times keywords associated with articles that mention AI over time. For example, \textit{chess} emerges most strongly in the late 1990s, after Deep Blue beats Kasparov.}
\label{fig:keywords}
\end{figure*}

\subsection{R3: What kinds of ideas are associated with AI, and how have they changed?}

The field of AI has changed enormously since 1986. What kinds of ideas did people associate with AI in the past, and how have these ideas changed in the present? 

To find out, we investigate the keywords most associated with AI-related articles from different time periods. We gather these keywords---for example, ``space'' or ``world politics'' or ``driverless vehicles''---from the New York Times API. We then group all New York Times articles into six five-year intervals between 1986 and 2016, and compute the mutual information (MI) between keyword counts and AI articles within each time period. For example, the keyword ``space'' might appear 80 times across all articles and 25 times in association with AI-related articles between 1986 and 1990, producing high MI with AI for that time period. This gives us a measure of the keywords most associated with AI articles over time. We then look across time periods and record themes in how these keywords change.

We present a sample of the keywords most strongly associated with AI for each time period in Figure \ref{fig:keywords}. Each keyword in the sample is among the 50 most related for that period and applies to at least two AI articles in the corpus. Some keywords (e.g., ``robot'') are common across all periods, and we did not include these in the sample. Other keywords (e.g., ``computer games'') remain strongly related to AI after the first time period in which they appear. 

The change in AI-associated keywords across time is revealing. From the concept of \textit{space weapons} in 1986:
\begin{quote}
\small
Real-time parallel processing may be the computational key to the creation of artificial intelligence, and conceivably to such functions as the control of President Reagan's Strategic Defensive Initiative, or Star Wars, program.
\normalsize
\end{quote}
To \textit{chess} in 1997:
\begin{quote}
\small
Even before the world chess champion Garry Kasparov faced the computer Deep Blue yesterday, pundits were calling the rematch another milestone in the inexorable advance of artificial intelligence.
\normalsize
\end{quote}
To \textit{search engines} in 2006:
\begin{quote}
\small
Accoona is a search engine that uses a heavy dose of artificial intelligence to find results that Google may miss.
\normalsize
\end{quote}
To \textit{driverless vehicles} in 2016: 
\begin{quote}
\small
United States vehicle safety regulators have said the artificial intelligence system piloting a self-driving Google car could be considered the driver under federal law.
\normalsize
\end{quote}

We also observe how the association of AI with individual keywords changes across time. For example, the association between AI and \textit{science fiction}, while present across all periods, peaks in the early 1990s. 



\begin{figure*}[!t]
\centering
\includegraphics[width=2.0\columnwidth]{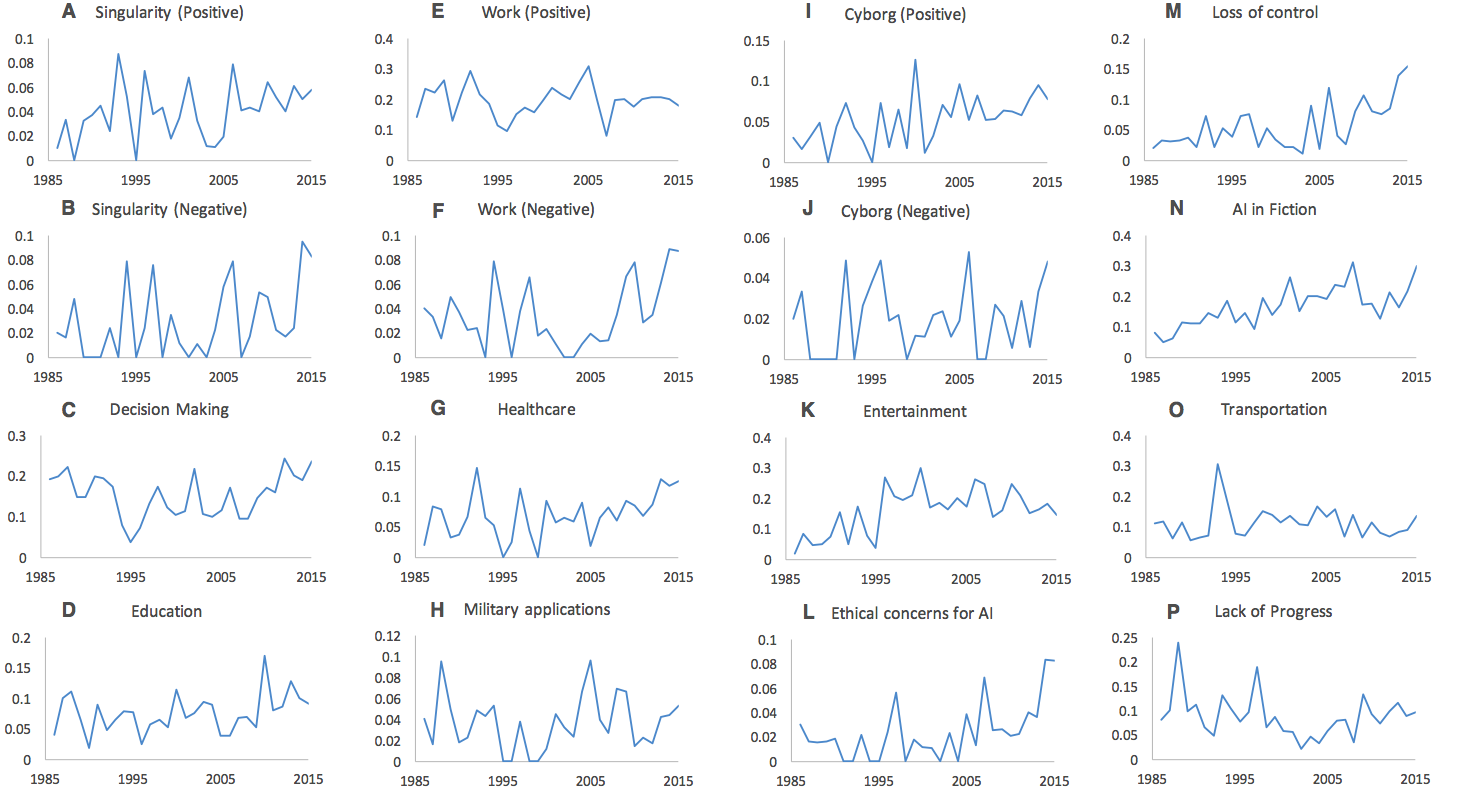}
\caption{Hopes and concerns from 1986 to 2016. In recent years, we see an increase in concern that humanity will lose of control of AI, and hope for the beneficial impact of AI on healthcare. The y-axis measures the percentage of AI articles that mention a specific hope or concern.}
\label{fig:index}
\end{figure*}

\subsection{R4: How have public hopes and concerns about AI changed over time?}

Beyond keywords, we studied indicators for fine-grained set of hopes and concerns related to AI. Here we examine how these ideas have evolved over time.

To this end, we examine all paragraphs tagged with AI hopes and concerns. We aggregate each of these paragraphs by article, and then by year. We consider an article as expressing a given hope or concern if it contains at least one paragraph that the crowd labeled with that concept. This gives us data that count the total number of times each AI hope and concern is expressed per article per year. We normalize these data by the total number of AI-related articles published per year, to arrive at a yearly percentage of AI-related articles that discuss each hope and concern.

We present the resulting trends in Figure \ref{fig:index}. While some data are sparse, we observe several clear upward trends. The fear of \textit{loss of control}, for example, has become far more common in recent years---more than triple what it was as a percentage of AI articles in the 1980s (Figure 3M). For example, in one article from 2009: 
\begin{quote}
\small
Impressed and alarmed by advances in artificial intelligence, a group of computer scientists is debating whether there should be limits on research that might lead to \textit{loss of human control} over computer-based systems that carry a growing share of society's workload, from waging war to chatting with customers on the phone.
\normalsize
\end{quote}
\textit{Ethical concerns} for AI have also become more common, driven in part by similar existential worries (Figure 3L). For example, in an article from 2015: 
\begin{quote}
\small
Two main problems with artificial intelligence lead people like Mr. Musk and Mr. Hawking to worry. The first, more near-future fear, is that we are starting to create machines that can make decisions like humans, but these machines \textit{don't have morality} and likely never will.
\normalsize
\end{quote}
These trends suggest an increase in public belief that we may soon be capable of building dangerous AI systems.

From a more positive standpoint, AI hopes for \textit{healthcare} have also trended upwards (Figure 3G). One strong theme is AI systems that care for patients. From 2003:
\begin{quote}
\small
For patients with more advanced cases, the researchers held out the possibility of systems that use artificial intelligence techniques to determine whether a person \textit{has remembered to drink fluids} during the day.
\normalsize
\end{quote}
Another driver of this trend is systems that can diagnose patients, or bioinformatics to cure disease. From 2013: 
\begin{quote}
\small
After Watson beat the best human Jeopardy champions in 2011, its artificial intelligence technology was directed toward new challenges, like assisting doctors in \textit{making diagnoses} in a research project at the Cleveland Clinic.
\normalsize
\end{quote}

In contrast, concerns over \textit{lack of progress} have decreased over time, despite a recent uptick (Figure 3P). This concern reached its high in 1988, at the start of the AI winter: 
\begin{quote}
\small
The artificial intelligence industry in general has been going through a \textit{retrenchment}, with \textit{setbacks} stemming from its failure to live up to its promises of making machines that can recognize objects or reason like a human.
\normalsize
\end{quote}
Intriguingly, many articles labeled with this concern in recent years draw reference to the past---a kind of meta-discussion about the \textit{lack of progress} concern itself. 

Among the remainder of the trends, a positive view of the impact of AI on human \textit{work} has become less common, while a negative view has increased sharply in recent years (Figure 3E-F). AI for \textit{education} has grown over time (Figure 3D), as has a positive view of \textit{merging} with AI (Figure 3I) and the role of AI in \textit{fiction} (Figure 3N).

\begin{figure}[!t]
\centering
\includegraphics[width=1.0\columnwidth]{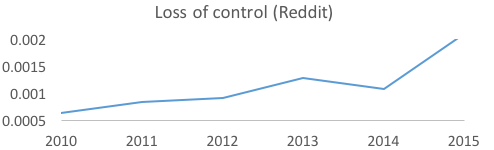}
\caption{We validated the increasing concern in \textit{loss of control} on Reddit data. The y-axis measures the percentage of AI-related comments that mention loss of control of AI.}
\label{fig:replicate}
\end{figure}

\section{News Articles from 1956 to 1986}

The New York Times provides full text for articles published after 1986, but article \textit{abstracts} (short descriptions of article content) are available over a much longer time period. To extend our results, we collected a dataset of all abstracts published between 1956 (the year of the first AI workshop at Dartmouth) and 1986.

Articles that mention AI are less common over this earlier period, with only 40 abstracts that reference AI (the first appears in 1977) and 247 that mention robots. These data are too sparse to extend our earlier analyses to 1956, but we have manually annotated each abstract with topic keywords to observe themes over time.  

In the 1950s, robots are most associated with military applications and especially missiles, e.g., ``the guided missile---the almost human robot of the skies.'' The 1960s and 70s strongly emphasize space, as in ``a ten-pound robot was shot into orbit today.'' Interest in AI picks up considerably in the early 1980s, where we see the first article that worries AI will negatively impact human jobs, the first reported death via robot, ``a factory worker killed by a robot arm,'' and the first mention of AI in healthcare, ``a robot to prepare meals and perform other chores for quadriplegics.''

\section{External validity}

Do the trends we have discovered in the New York Times generalize to the public at large? While this question is difficult to answer directly, we have replicated one of our primary findings on 5 years of public posts from Reddit, a popular online community with a diverse set of users. 

Concretely, we train a classifier to predict the presence of \textit{loss of control} in paragraphs about AI using our annotated data from the New York Times. We then apply this classifier to posts made by Reddit users. We use a logistic regression model based on TF-IDF features and threshold the positive class probability at 0.9. In validation, we observe precision of 0.8 on a sample of 100 Reddit posts annotated with ground truth. Finally, we apply this classifier to every post that mentioned ``artificial intelligence'' from 2010 to 2015.

We present the resulting trend in Figure \ref{fig:replicate}, which mirrors Figure 3M over the same time period. Broadly, this replication suggests that attitudes among Reddit users shift in line with what we see in the New York Times, providing some evidence for the external validity of our findings.

\section{Related Work}

Others have discussed the impact of artificial intelligence on society and the range of future outcomes \cite{ai-concerns}. These discussions are in part driven by a need to address public concerns about AI---our work is the first to quantify such concerns through direct analysis. The set of indicators we have introduced will be useful in framing future discussions, such as those ongoing in the One Hundred Year Study of Artificial Intelligence \cite{100years}.  

Public opinion polls have similarly measured topics relevant to AI. While such polls are recent (and not conducted over time), they support our findings, showing greater levels of optimism than pessimism about AI, but increasing existential fear and worry about jobs \cite{bsa-poll,cbs-poll}. Future polls might allow us to directly measure public opinion on the set of measures we have studied. 

Beyond artificial intelligence, other work has mined cultural perspectives from text corpora over long time periods. For example, by analyzing 200 years of data from Google Books, it is possible to quantify the adoption of new technologies or changes in psychological attitudes through linguistic patterns \cite{google-culture,psych-culture}. Using music, others have quantified changes in artistic style over a 40 year period \cite{music-over-time}. We use crowdsourced annotations to extend the limits of what is possible under these kinds of quantitative analyses.


News and social media offer a powerful reflection of public attitudes over time. For example, by analyzing such data, it is possible to predict cultural events such as revolutions \cite{future1,future2}, or examine public opinion on same-sex marriage \cite{marriage}. Here we use such data to discover and validate similar trends in the public perception of artificial intelligence. 

Finally, crowdsourcing is a powerful tool for enabling new kinds of quantitative analyses. For example, it is possible to crowdsource lexicons of words to answer novel research questions \cite{empath}, or leverage crowds to bootstrap classifiers that can then be applied to much larger corpora \cite{politeness,dogmatism}. Here we use crowds to identify themes in articles that would be difficult to analyze under fully automated approaches.

\section{Conclusion}
We present a set of indicators that capture levels of engagement, general sentiment, and hopes and concerns for the future of artificial intelligence over time. We then validate these impression indicators by studying trends in 30 years of articles from the New York Times. We find that discussion of AI has increased sharply since 2009 and has been consistently more optimistic than pessimistic. However, many specific concerns, such as the fear of loss of control of AI, have been increasing in recent years. 

\bibliographystyle{aaai}
\bibliography{ref}

\begin{thebibliography}{}

\bibitem[\protect\citeauthoryear{{60 Minutes}}{2016}]{cbs-poll}
{60 Minutes}.
\newblock 2016.
\newblock 60 minutes poll: Artificial intelligence.

\bibitem[\protect\citeauthoryear{BSA}{2015}]{bsa-poll}
BSA.
\newblock 2015.
\newblock One in three believe that the rise of artificial intelligence is a
  threat to humanity.

\bibitem[\protect\citeauthoryear{Danescu-Niculescu-Mizil \bgroup et
  al\mbox.\egroup }{2013}]{politeness}
Danescu-Niculescu-Mizil, C.; Sudhof, M.; Jurafsky, D.; Leskovec, J.; and Potts,
  C.
\newblock 2013.
\newblock A computational approach to politeness with application to social
  factors.
\newblock {\em arXiv preprint arXiv:1306.6078}.

\bibitem[\protect\citeauthoryear{Dietterich and Horvitz}{2015}]{ai-concerns}
Dietterich, T.~G., and Horvitz, E.~J.
\newblock 2015.
\newblock Rise of concerns about {AI}: reflections and directions.
\newblock {\em Communications of the ACM} 58(10):38--40.

\bibitem[\protect\citeauthoryear{Doroudi \bgroup et al\mbox.\egroup
  }{2016}]{examples}
Doroudi, S.; Kamar, E.; Brunskill, E.; and Horvitz, E.
\newblock 2016.
\newblock Toward a learning science for complex crowdsourcing tasks.
\newblock In {\em Proceedings of the 2016 CHI Conference on Human Factors in
  Computing Systems},  2623--2634.
\newblock ACM.

\bibitem[\protect\citeauthoryear{Fast and Horvitz}{2016}]{dogmatism}
Fast, E., and Horvitz, E.
\newblock 2016.
\newblock Identifying dogmatism in social media: Signals and models.
\newblock In {\em Proceedings of EMNLP 2016}.

\bibitem[\protect\citeauthoryear{Fast, Chen, and Bernstein}{2016}]{empath}
Fast, E.; Chen, B.; and Bernstein, M.~S.
\newblock 2016.
\newblock Empath: Understanding topic signals in large-scale text.
\newblock In {\em Proceedings of the 2016 CHI Conference on Human Factors in
  Computing Systems},  4647--4657.
\newblock ACM.

\bibitem[\protect\citeauthoryear{Greenfield}{2013}]{psych-culture}
Greenfield, P.~M.
\newblock 2013.
\newblock The changing psychology of culture from 1800 through 2000.
\newblock {\em Psychological science} 24(9):1722--1731.

\bibitem[\protect\citeauthoryear{Guardian}{2014}]{elon-regulatory}
Guardian, T.
\newblock 2014.
\newblock Elon {Musk}: Artificial intelligence is our biggest existential
  threat.

\bibitem[\protect\citeauthoryear{Horvitz and Selman}{2009}]{asilomar}
Horvitz, E., and Selman, B.
\newblock 2009.
\newblock Interim report from the panel chairs, {AAAI} presidential panel on
  long-term {AI} futures.
\newblock {\em Association for the Advancement of Artificial Intelligence,
  August 2009}.

\bibitem[\protect\citeauthoryear{Leetaru}{2011}]{future2}
Leetaru, K.
\newblock 2011.
\newblock Culturomics 2.0: Forecasting large-scale human behavior using global
  news media tone in time and space.
\newblock {\em First Monday} 16(9).

\bibitem[\protect\citeauthoryear{Michel \bgroup et al\mbox.\egroup
  }{2011}]{google-culture}
Michel, J.-B.; Shen, Y.~K.; Aiden, A.~P.; Veres, A.; Gray, M.~K.; Pickett,
  J.~P.; Hoiberg, D.; Clancy, D.; Norvig, P.; Orwant, J.; et~al.
\newblock 2011.
\newblock Quantitative analysis of culture using millions of digitized books.
\newblock {\em Science} 331(6014):176--182.

\bibitem[\protect\citeauthoryear{ProPublica}{2016}]{machine-bias}
ProPublica.
\newblock 2016.
\newblock Machine bias in criminal sentencing.

\bibitem[\protect\citeauthoryear{Radinsky and Horvitz}{2013}]{future1}
Radinsky, K., and Horvitz, E.
\newblock 2013.
\newblock Mining the web to predict future events.
\newblock In {\em Proceedings of the sixth ACM international conference on Web
  search and data mining},  255--264.
\newblock ACM.

\bibitem[\protect\citeauthoryear{Salehi, Irani, and Bernstein}{2015}]{dynamo}
Salehi, N.; Irani, L.~C.; and Bernstein, M.~S.
\newblock 2015.
\newblock We are {Dynamo}: Overcoming stalling and friction in collective
  action for crowd workers.
\newblock {\em Proceedings of the 33rd Annual ACM Conference on Human Factors
  in Computing Systems}  1621--1630.

\bibitem[\protect\citeauthoryear{Serr{\`a} \bgroup et al\mbox.\egroup
  }{2012}]{music-over-time}
Serr{\`a}, J.; Corral, {\'A}.; Bogu{\~n}{\'a}, M.; Haro, M.; and Arcos, J.~L.
\newblock 2012.
\newblock Measuring the evolution of contemporary western popular music.
\newblock {\em Nature Scientific reports}.

\bibitem[\protect\citeauthoryear{Sheng, Provost, and
  Ipeirotis}{2008}]{get-another-label}
Sheng, V.~S.; Provost, F.; and Ipeirotis, P.~G.
\newblock 2008.
\newblock Get another label? {Improving} data quality and data mining using
  multiple, noisy labelers.
\newblock {\em Proceedings of the 14th ACM SIGKDD international conference on
  Knowledge discovery and data mining}  614--622.

\bibitem[\protect\citeauthoryear{{Stanford University}}{2014}]{100years}
{Stanford University}.
\newblock 2014.
\newblock One hundred year study on artificial intelligence.

\bibitem[\protect\citeauthoryear{{Stone, P. et al.}}{2016}]{ai100}
{Stone, P. et al.}
\newblock 2016.
\newblock Artificial intelligence and life in 2030.
\newblock {\em One Hundred Year Study on Artificial Intelligence: Report of the
  2015-2016 Study Panel}.

\bibitem[\protect\citeauthoryear{Zhang and Counts}{2015}]{marriage}
Zhang, A.~X., and Counts, S.
\newblock 2015.
\newblock Modeling ideology and predicting policy change with social media:
  Case of same-sex marriage.
\newblock In {\em Proceedings of the 33rd Annual ACM Conference on Human
  Factors in Computing Systems},  2603--2612.
\newblock ACM.

\end{thebibliography}

\end{document}